\documentclass[10pt, conference, letterpaper]{IEEEtran}
\usepackage{amsmath,amsfonts}
\usepackage{amssymb}
\usepackage{algorithmic}
\usepackage{algorithm}
\usepackage{array}
\usepackage[caption=false,font=normalsize,labelfont=sf,textfont=sf]{subfig}
\usepackage{textcomp}
\usepackage{stfloats}
\usepackage{url}
\usepackage{verbatim}
\usepackage{graphicx}
\usepackage{multirow}
\usepackage{makecell}
\usepackage{cite}
\usepackage{arydshln}
\usepackage{booktabs}
\def\BibTeX{{\rm B\kern-.05em{\sc i\kern-.025em b}\kern-.08em
T\kern-.1667em\lower.7ex\hbox{E}\kern-.125emX}}

\usepackage[colorlinks=true, allcolors=blue]{hyperref}
\makeatletter

\newcommand{\Rmnum}[1]{\expandafter\@slowromancap\romannumeral #1@}
\makeatother

\title{Dual-Phase Federated Deep Unlearning via Weight-Aware Rollback and Reconstruction}

\author{
Changjun~Zhou\textsuperscript{*},
Jintao~Zheng\textsuperscript{*},
Leyou~Yang\textsuperscript{†},
Pengfei~Wang\textsuperscript{‡}\\
\textsuperscript{*}School of Computer Science and Technology, Zhejiang Normal University, Jinhua, China \\
\textsuperscript{†}School of Software, Nanjing University of Information Science and Technology, Nanjing, China \\
\textsuperscript{‡} School of Computer Science and Technology, Dalian University of Technology, Dalian, China \\
E-mails: \{zhouchangjun, 00taotao7\}@zjnu.edu.cn, yangleyou@163.com, wangpf@dlut.edu.cn
\thanks{This work was supported by National Natural Science Foundation of China under grants 62272418, 72342013, 72371054, 62202080 and 62320106006, Zhejiang Provincial Natural Science Foundation Major Project under grant~LD24F020004, the Major Open Project of Key Laboratory for Advanced Design and Intelligent Computing of the Ministry of Education under grant~ADIC2023ZD001, the China Postdoctoral Science Foundation under grant~2023M733354, the Science and Technology Project of Liaoning Province under grant~2023JH1/10400083, the Dalian Science and Technology Talent Innovation Support Plan for Outstanding Young Scholars under grant~2023RY023, Xiaomi Young Talents Program, and the Fundamental Research Funds for the Central Universities under grant~DUT25GF206. 
}
\thanks{(\textit{Corresponding author: Pengfei Wang})}
}

\begin{document}

\IEEEoverridecommandlockouts

\maketitle

\begin{abstract}
Federated Unlearning~(FUL) focuses on client data and computing power to offer a privacy-preserving solution. However, high computational demands, complex incentive mechanisms, and disparities in client-side computing power often lead to long waiting time and high costs. To address these challenges, many existing methods rely on server-side knowledge distillation that solely removes the updates of the target client, overlooking the privacy embedded in the contributions of other clients, which can lead to privacy leakage. In this work, we introduce DPUL, a novel server-side unlearning method that deeply unlearns all influential weights to prevent privacy pitfalls. 
Our approach comprises three components: $(i)$~identifying high-weight parameters by filtering client update magnitudes, and rolling them back to ensure deep removal, $(ii)$~leveraging the variational autoencoder~(VAE) to reconstruct and eliminate low-weight parameters, $(iii)$~utilizing a projection-based technique to recover the model. Experimental results on four datasets demonstrate that DPUL surpasses state-of-the-art baselines, providing a $1\%–5\%$ improvement in accuracy and up to $12\times $ reduction in time cost. 
\end{abstract}
\begin{IEEEkeywords}
Federated unlearning, large model, deep unlearning, data trading.
\end{IEEEkeywords}

\section{Introduction}


Federated Unlearning (FUL) is an emerging approach designed to remove the influence of specific parameters within Federated Learning (FL) paradigms~\cite{FU_background,FUL_background2,FL_infocom_2025}.
The primary objective of FUL is to eliminate particular contributions~\cite{GDPR,CCPA} in the model parameters while maintaining accuracy levels comparable to those before unlearning. The conventional method is to retrain the model from scratch~\cite{FUL_infocom_2025}. However, due to the significant time and computational costs involved, the retraining approach is usually used as a baseline for comparison rather than as a practical solution~\cite{FUL_BK3}. 
FUL has broad applications in domains such as healthcare and finance, where data privacy restrictions prevent direct data sharing~\cite{application,FUL_background}.

\begin{figure}
    \centering
    \includegraphics[width=\linewidth]{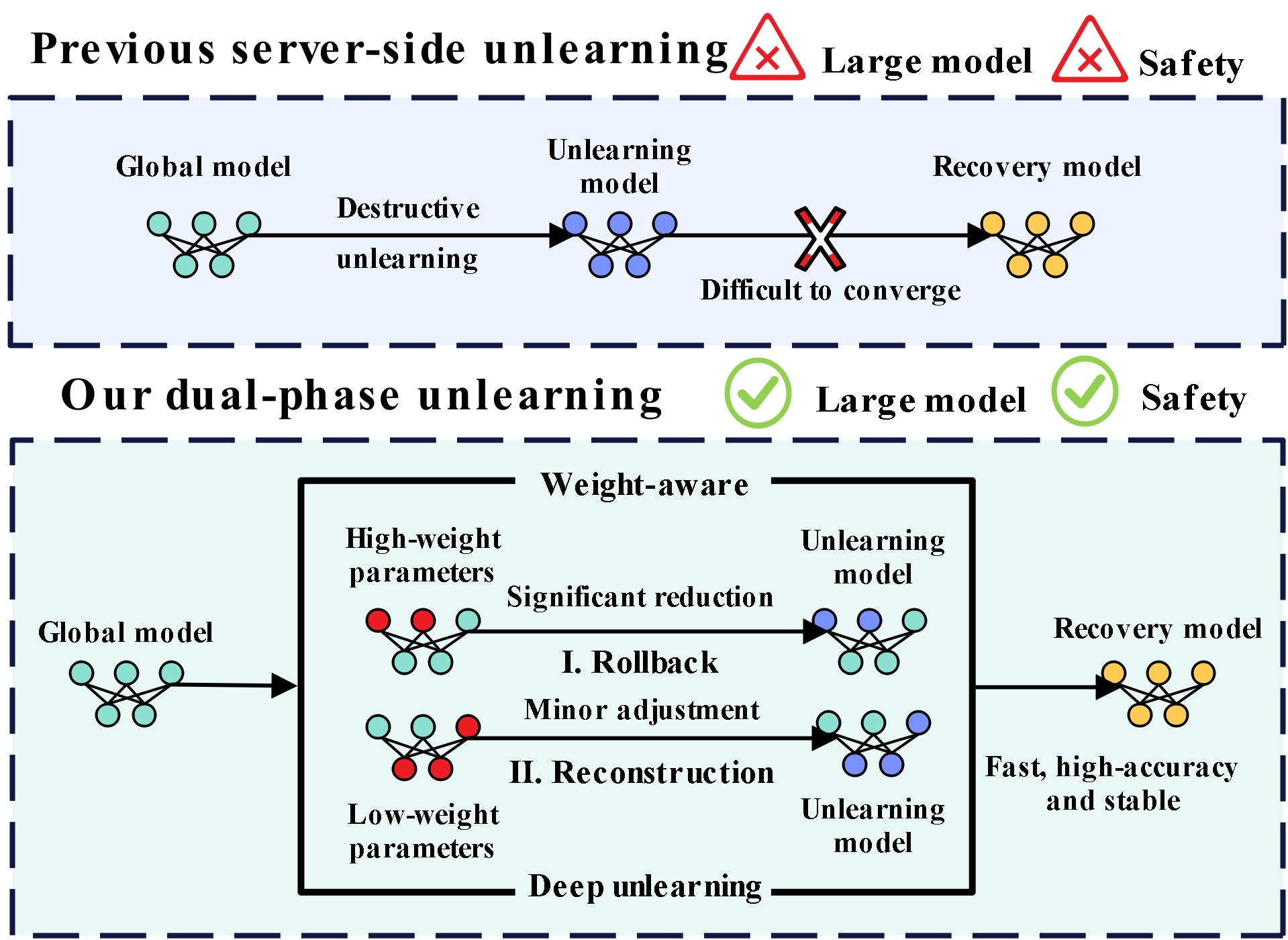}
    \caption{\textbf{The comparison between the previous method and DPUL.}}
    \label{fig:intro}
\vspace{-0.3cm}
\end{figure}

Current approaches to FUL rely on clients, a process that is both time-consuming and costly~\cite{Communication_efficient}. In large-scale model training, computational demands grow with the number of parameters~\cite{large_infocom_2025_2}. When servers rely on clients for unlearning, they often incur substantial computing costs and time~\cite{client_infocom_2025}. If original clients have already left, maintaining the accuracy of model recovery becomes a challenge~\cite{client_FL}.
Furthermore, the clients’ computing power varies, and the least capable client determines the unlearning duration, which inevitably extends the process.
Few methods enable unlearning solely on the server side, and they are insufficient in meeting the demands for fast and efficient unlearning~\cite{client_selection}.

Server-based unlearning methods, such as knowledge distillation, are the primary approaches to avoid long time and high cost~\cite{FD}. However, large-scale model recovery requires extensive distillation data and significant processing time~\cite{large_model_infocom_2025,FD_infocom_2025}. Additionally, only the target client’s updates are discarded, without accounting for residual privacy traces in other clients’ contributions, which may be inadvertently distilled and recovered into the model and lead to privacy leakage.

To address these challenges, we introduce the Dual-Phase Federated Deep Unlearning method~(DPUL), a novel approach designed for efficient and safe unlearning by deeply unlearning all influential weights. On the server side, our method first leverages the relationship between the target client’s and global parameters to filter out those with high-weight contributions. Then it applies the memory rollback method to process these parameters, ensuring precise unlearning. We utilize the reconstruction properties of $\beta$-VAE~\cite{beta-VAE} for lower-weight contributions, enabling the model to unlearn less significant influences. Finally, a small dataset and projection method are incorporated to facilitate recovery, helping to maintain model performance while ensuring successful unlearning. Figure \ref{fig:intro} compares the previous method and DPUL.

Our method offers several significant advantages over existing approaches. First, it operates without client involvement, eliminating the drawbacks associated with client-based unlearning, such as high costs, prolonged training time, and potential instability in results. Second, since unlearning occurs entirely on the server side, the process is highly stable and swift, ensuring reliable performance. Finally, our approach reconstructs all parameters, eliminating the target client’s parameter-specific contribution while maintaining overall model accuracy and avoiding privacy leakage.

Despite its advantages, our method also presents several challenges. First, screening client high-weight contribution parameters is a complex task, especially given the large number of parameters in models. To address it, we will conduct a statistical analysis of parameter updates. Suppose the amplitude of the parameter update from the target client exceeds the average too much. In that case, we classify the parameter as a client high-weight contribution parameter and revert it to the earliest low-weight state across all training rounds. Second, training VAE unlearning networks is crucial, as different processes yield varying results. Identifying the most suitable training method is essential to ensure that reconstructed parameters deviate maximally from the original model while maintaining high accuracy. To address it,  we introduce a multi-head training mechanism, which segments model parameters and trains multiple unlearning networks in parallel, significantly improving the unlearning process. Finally, the refined model must maintain high-precision performance, ensuring its reliability and effectiveness post-unlearning. To address it, we leverage accuracy metrics to match similar global models. We employ the projection technique~\cite{federaser} for acceleration, assisting in model recovery and ensuring performance stability post-unlearning.



The contributions of our paper are as follows: 
\begin{itemize}
\item[$\bullet$] We propose an innovative approach for filtering client contributions by leveraging the magnitude of client model updates and global model to determine whether a parameter qualifies as a client high-weight contribution, which can filter high-weight contribution parameters.
\item[$\bullet$] We propose an unlearning network approach by leveraging the Variational Autoencoder~(VAE) neural network’s capability to identify and remove client parameter contributions, eliminating low-weight contributions. 
\item[$\bullet$] We introduce a multi-head training mechanism by partitioning model parameters into multiple segments and assigning an equal number of unlearning networks to each, significantly improving the VAE performance. 
\item[$\bullet$] We conduct extensive experiments across four datasets and network architectures to verify the performance of our proposed DPUL method. 
Experimental results indicate that our approach achieves a $12\times$ improvement in efficiency compared to retraining and delivers a $1\%–5\%$ improvement in accuracy.
\end{itemize}


\section{Related Work}
FUL has many practical applications, including the removal of sensitive data, the mitigation of backdoor attacks, and the elimination of low-quality data in federated training~\cite{FU_background2,Asynchronous_FUL,privacy_FL}. Traditional retraining methods often struggle with slow recovery speeds and high costs. To address these challenges, FURR~\cite{RR} introduces fast retraining and employs the Hessian matrix approximation to accelerate the training process and facilitate rapid unlearning. Similarly, FedEraser~\cite{federaser} leverages server-stored historical updates to correct retraining bias and improve efficiency. UPGA~\cite{GA}, on the other hand, utilizes gradient ascent for privacy unlearning, followed by a restoration phase enhanced through small dataset boosting. However, these experimental approaches require clients’ participation, which entails significant costs and time investments.
Furthermore, FUCP~\cite{fucp} employs model pruning to facilitate category unlearning, while FAST~\cite{guo2023fast} utilizes a historical update-erasure method to eliminate malicious clients. SIFU~\cite{sifu} adopts gradient ascent to remove low-quality data, and F2UL~\cite{F2UL} allocates distinct models to clients based on their data quality to promote fairness. RFUL~\cite{Fast_recovery} utilizes the one-vs-rest network to enable unlearning clients that have exited to recover their previous contributions quickly. CGKD~\cite{clip} aims to use a large CLIP model to assist in model recovery. FedCF~\cite{mutil_FUL_local} utilizes a reconstruction model to accelerate the recovery of the remaining client contributions, providing feasibility for multiple clients to unlearn simultaneously. These approaches extend the application of FUL beyond client privacy, addressing broader challenges within FL.

As models develop, they become more complex and require greater computational resources. Consequently, the costs associated with client-side privacy unlearning methods have surged dramatically. FUKD~\cite{KD} seeks to leverage extensive data and prolonged processing on the server side to distill and restore models. However, it is no longer viable for scenarios requiring unlearning large models. To address this limitation, we propose the DPUL method, which directly utilizes server resources for FUL, mitigating excessive costs and reducing time consumption.



\section{Method}


\subsection{Overview}

To begin, we provide a comprehensive overview of our proposed method. We first define the set of clients participating in FL as $\mathcal{N} = \{1, 2, ..., N\}$ and outline their respective datasets, $\mathcal{D}_i = \{ (x_k,y_k)\mid i \in \mathcal{N}, k = 1, 2, ...\}$, where $x_k \in \mathbb{R}^d$ represents a d-dimensional input vector (e.g., images) and $y_k$ denotes the corresponding ground truth label. We introduce the global model as $\mathcal{M}$, the global cumulative updates as $\bigtriangleup \mathcal{M}$, and the cumulative updates specific to the target client as $\bigtriangleup \mathcal{U}$. When a target client requests to unlearn its data, the server initiates our method with the memory rollback method. We first process the historical parameters of $\mathcal{M}$ and eliminate high-weight parameters associated with the target client, thereby ensuring data privacy and yielding the globally processed model defined as $\mathcal{M}^{'}$. Subsequently, we address low-weight parameters that may retain limited private information. To achieve this, we leverage the reconstruction unlearning method. We use the reconstruction capabilities of a Variational Autoencoder (VAE) to obtain the final unlearning model, $\mathcal{M}^u$. Finally, we use the projection method with the small dataset to recover the effectiveness of the unlearning model.

\begin{figure}[t!]
    \centering
    \includegraphics[width=\linewidth]{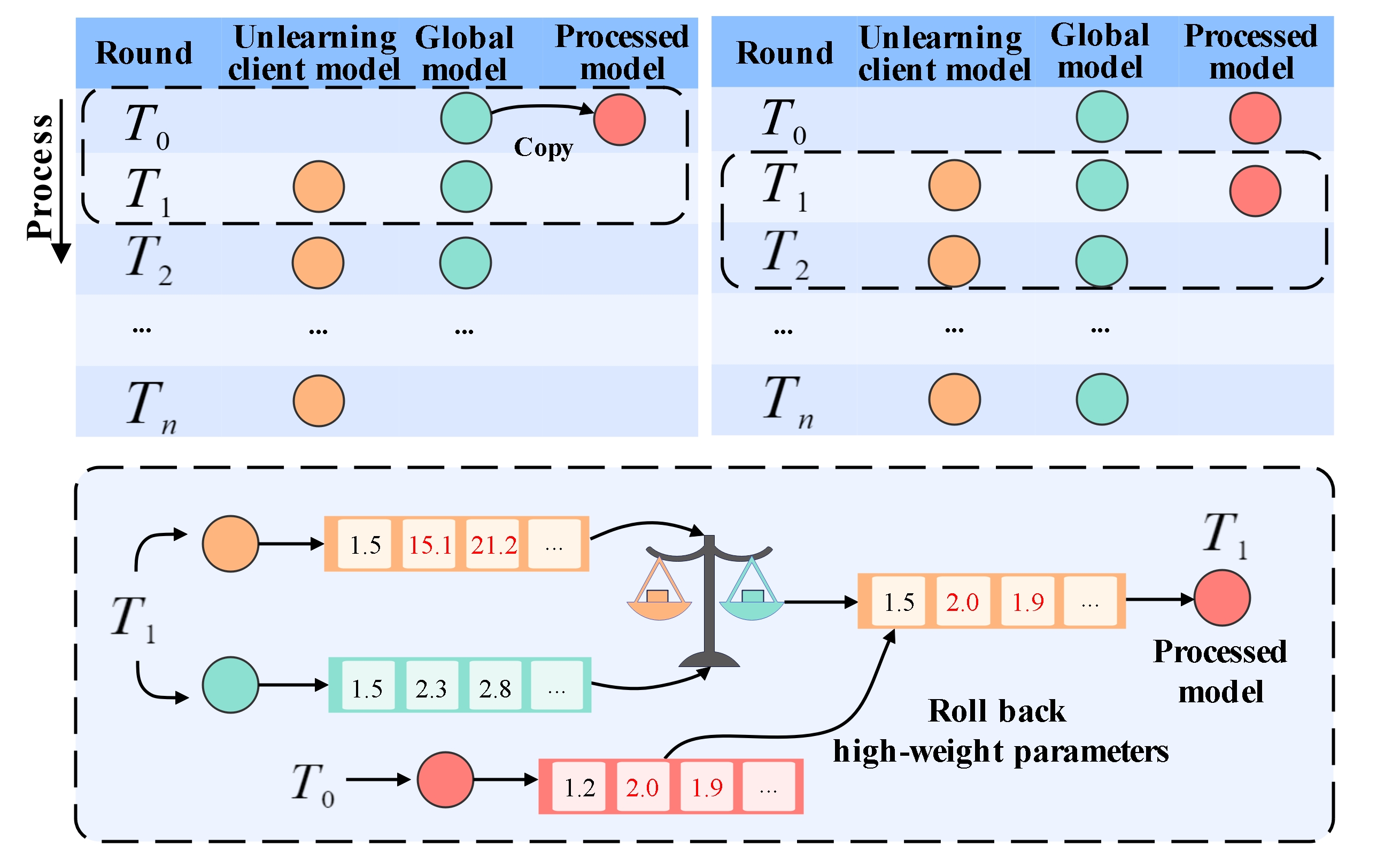}
    \caption{\textbf{The overall workflow of the memory rollback.}}
    \label{fig:Memory regression}
\end{figure}

During the memory rollback phase, we assess the target client’s cumulative update magnitude $\bigtriangleup \mathcal{U}$. If the magnitude is deemed excessive, we categorize that parameter as a high-weight parameter potentially harboring sensitive privacy contribution, and we revert it to the nearest parameter state that lacks significant contribution. Finally, we obtain the global processed model $\mathcal{M}^{'}$, as depicted in Figure \ref{fig:Memory regression}.

We treat low-weight contribution parameters as potentially harboring minor secondary privacy concerns. We then train a VAE neural network using both the global historical parameters $\mathcal{M}$ and the preprocessed parameters $\mathcal{M}^{'}$, aiming to learn how to unlearn the contributions of the target client. This approach enables us to reconstruct all network model parameters in a manner that effectively erases the target client’s influence. To expedite VAE training, we employ a slice training method by segmenting both $\mathcal{M}$ and $\mathcal{M}^{'}$, and constructing VAE networks $\mathcal{V}$ with an equal number of slices for parallel training to facilitate rapid convergence. After training, we use the VAE networks $\mathcal{V}$ to obtain the unlearning model $\mathcal{M}^u$.


Finally, during the recovery phase, we employ a small-scale dataset defined as $\mathcal{D}_{un}$ to assist in the recovery process via the projected boost method, ensuring that high performance is maintained while effectively unlearning all features associated with the target client.
Below, we describe our algorithms for the memory rollback, the reconstruction unlearning, and the projected boost recovery.



\subsection{Memory Rollback}
First, the server stores the cumulative updates of the global model, $\bigtriangleup \mathcal{M}$, and the cumulative updates from the client, $\bigtriangleup \mathcal{U}$. We then process the global model’s cumulative updates to ensure they do not include any parameter contributions from the client. For each round $t$, we have
\begin{equation}
\begin{aligned}
        \bigtriangleup \mathcal{M}^u &= \bigtriangleup \mathcal{M} \setminus  \bigtriangleup \mathcal{U} \\
        & = \frac{1}{1-p_u}(\bigtriangleup \mathcal{M} - p_u\bigtriangleup \mathcal{U})
\end{aligned}
\end{equation}
where $\bigtriangleup \mathcal{M}^u$ is the global cumulative updates except for the unlearning client, $p_i = \frac{\left | D_i \right | }{\sum_{1}^{N} \left | D_i \right | }$ represents the weight value where $\left | D_i \right |$ is the size of the client datasets.

For the $t$-th round, we evaluate each parameter update $\bigtriangleup \mathcal{U}^t_i$ of the client’s model, where $i$ denotes the $i$-th parameter. If  $\frac{\left|\bigtriangleup \mathcal{U}^t_i\right|}{\lambda} > \left |{\bigtriangleup \mathcal{M}^{u}}^t_i\right | $, we classify this parameter as a high-weight contribution where $\lambda$ is high-weight coefficient and $\left|\cdot \right|$ is absolute value. Consequently, we roll back the subsequent updates of this parameter $\mathcal{M}_i$ to the nearest low-weight contribution parameter. If this occurs in the first round, we remove the unlearning client’s parameters. We obtain the processed global model $\mathcal{M}^{'}$, excluding the high-weight contribution parameters from the client’s updates. The specific details are presented in Algorithm \ref{Memory regression}.

\begin{figure}
	\renewcommand{\algorithmicrequire}{\textbf{Input:}}
	\renewcommand{\algorithmicensure}{\textbf{Output:}}
    \begin{algorithm}[H]
        \caption{Memory rollback}
        \label{Memory regression}
        \begin{algorithmic}[1]
            \REQUIRE Global models $\mathcal{M}$,  Global model update parameters $\bigtriangleup \mathcal{M}$, Unlearning client update parameters $\bigtriangleup \mathcal{U}$, High weight coefficient $\lambda$.    
            \ENSURE Processing global models  $\mathcal{M} ^{'}$.     
            \STATE{Set $\mathcal{M}^{'}$ = $\mathcal{M}$ };
            \STATE{Set $\bigtriangleup \mathcal{M}^{u} = \bigtriangleup \mathcal{M} \setminus  \bigtriangleup \mathcal{U}$};
            \FOR{t =1, 2, 3, ..., T}
            \IF{t == 1}
            \STATE{${\mathcal{M}^{'}}^t = {\mathcal{M}^{'}}^t  - \bigtriangleup {\mathcal{M}^{u}}^t$};
            \STATE{continue};
            \ENDIF
            \FOR{$\bigtriangleup \mathcal{U}^t_i \in \bigtriangleup \mathcal{U}^t$ }

            \IF{$\frac{\left |\bigtriangleup \mathcal{U}^t_i\right |}{\lambda}  > \left |{\bigtriangleup \mathcal{M}^{u}}^t_i\right |  $}
            \FOR{j = t, t+1, ..., T}
            \STATE{${\mathcal{M}^{'}} _i^j= {\mathcal{M}^{'}}_i^{j - 1}$};
            \ENDFOR
            \ENDIF
            \ENDFOR
            \ENDFOR
            \RETURN{$\mathcal{M}^{'}$}
        \end{algorithmic}
    \end{algorithm}
\vspace{-0.7cm}
\end{figure}

\subsection{Reconstruction Unlearning}
Initially, we partition both the processed model $\mathcal{M}^{'}$ and the global model $\mathcal{M}$ into slices, thereby yielding a set of corresponding segments that serve as inputs for dedicated VAE networks. We have
\begin{equation}
\begin{aligned}
    &Split(\mathcal{M}) \to \{\mathcal{M}(1), \mathcal{M}(2), ..., \mathcal{M}(I)\}\\
    &Split(\mathcal{M}^{'}) \to \{\mathcal{M}^{'}(1), \mathcal{M}^{'}(2), ..., \mathcal{M}^{'}(I)\}
\end{aligned}
\label{slice}
\end{equation}
where $\mathcal{M}(i)$ and $\mathcal{M}^{'}(i)$ represent the $i$-$th$ slice of parameters.

For the $i$-$th$ slice of parameters, we construct a $\beta$-VAE network, denoted as $V^i$. The input and output dimensions for $V^i$ are aligned with the size of the model parameters $\mathcal{M}(i)$. To build the dataset, we incorporate both the global model parameters $\mathcal{M}(i)$ and the processing parameters $\mathcal{M}^{'}(i)$, resulting in $\mathcal{D}(i) = \{(x_t = \mathcal{M}^t(i),{y_t = \mathcal{M}{'}}^t(i))\mid i = {1, 2, ..., I}\; ; \;t = 1, 2, ..., T - 1\}\}$, where $T$ represents the total number of rounds in FL. The training loss function and optimization problem are defined as

\begin{equation}
\begin{aligned}
    &\ell (V) = \mathbb{E}_{q(z|x)}[(y - \hat{y}(z))^2] + \beta D_{KL}(q(z|x) \| p(z)) \\
    &\min_{V^i\in \mathbb{R}^d} L_i(V^i) = \frac{1}{n_i}\sum_{k\in D(i)}\ell_i (V^i)\\
\end{aligned} 
\label{beta}
\end{equation}
where $q(z|x)$ represents approximate posterior, $y$ is target label, $\hat{y}(z)$ is output of the model, $\beta$ represents the hyperparameter, $p(z)$ indicates the prior distribution, $D_{KL}$ is $KL$ divergence which measures the difference between two probability distributions, $L_i(V^i)$ represents the loss function of $V^i$, $n_i$ presents the data size, and $\ell_i$ is logits output.



We then train each neural network $V^i$ using the dataset $\mathcal{D}(i)$ for $E$ rounds. Upon completing the training process, we utilize $V^i$ to derive the unlearning model $\mathcal{M}^u$, which is 

\begin{equation}
\begin{aligned}
    &\mathcal{M}^u(i) = V^i(M^T)\\
    &\mathcal{M}^u = Union(\mathcal{M}^u(1), \mathcal{M}^u(2),..., \mathcal{M}^u(I))
\end{aligned}
\end{equation}

Finally, we obtain the unlearning model $\mathcal{M}^u$. The specific details are provided in Algorithm \ref{Deep unlearning}.
\begin{figure}[htbp]
	\renewcommand{\algorithmicrequire}{\textbf{Input:}}
	\renewcommand{\algorithmicensure}{\textbf{Output:}}
    \begin{algorithm}[H]
        \caption{Reconstruction unlearning}
        \label{Deep unlearning}
        \begin{algorithmic}[1]
            \REQUIRE Global models $\mathcal{M}$, Processing global models $\mathcal{M}^{'}$, Small amount of datasets $D_s$.
            \ENSURE Unlearning model $\mathcal{M}^u$.     
            \FOR{$i = 1, 2, ..., I$}
            \FOR{$t = 1, 2, ..., T$}
            \STATE{$\mathcal{M}^t(i) \gets Split(\mathcal{M}^t)$ }
            \STATE{${\mathcal{M}^{'}}^t(i) \gets Split({\mathcal{M}^{'}}^t)$ }
            \ENDFOR
            \STATE{Create network $V^i$}
            \ENDFOR
            
            \FOR{$\mathcal{M}^t(i) \in  \mathcal{M}^t$ \textbf{in parallel}}
            \FOR{$Epoch = 1, 2, ..., E$}
            \FOR{$t = 1, 2, ..., T - 1$}
            \STATE{$Train(V^i, \mathcal{M}^t(i),{\mathcal{M}^{'}}^t(i))$};
            \ENDFOR
            \ENDFOR
            \ENDFOR
            \FOR{$\mathcal{M}^t(i) \in  \mathcal{M}^t$ \textbf{in parallel}}
            \STATE{$\mathcal{M}^u(i) \gets V^i(M^{T})$};
            \ENDFOR
            \STATE{$\mathcal{M}^u \gets Union(\mathcal{M}^u(1), \mathcal{M}^u(2), ...,\mathcal{M}^u(I))$};
            \RETURN{$\mathcal{M}^u$}
        \end{algorithmic}
    \end{algorithm}
\vspace{-0.7cm}
\end{figure}
\subsection{Projected Boost Recovery}
The server records the accuracy of each update round, denoted as $L_{acc} = \{Acc(\mathcal{M}^1), Acc(\mathcal{M}^2),..., Acc(\mathcal{M}^T)\}$, where $Acc(\mathcal{M}^t)$ represents the accuracy of the global model $\mathcal{M}^t$. When the server initiates unlearning recovery, the first step is to determine the projected round $t$, given by $t = \arg\min(\left | L_{acc}^t - Acc(\mathcal{M}^u) \right | )$, where $L_{acc}^t$ refers to the value of the $t$-$th$ element in $L_{acc}$. For computational convenience, we select the first round $t$ in which the accuracy $L_{acc}^t$ exceeds $Acc(\mathcal{M}^u)$.

We utilize the dataset $D_{b}$ and apply projection methods to expedite the recovery of the unlearning model. Initially, we enhance the unlearning model $\mathcal{M}^u$ using the dataset $D_{b}$, resulting in the update $\bigtriangleup \mathcal{M}^{u}$. Specifically, we define the optimization problem as   
\begin{equation}
    \min_{\mathcal{M}^u\in \mathbb{R}^d}L_b(\mathcal{M}^u) = \frac{1}{n_{b}}\sum_{i\in D_{b}}{\ell_b}
\end{equation}
and $\bigtriangleup \mathcal{M}^{u} = \eta \bigtriangledown L_b(\mathcal{M}^u )$, where $\eta$ represents the learning rate and $ \bigtriangledown L_b$ is gradient of loss function. We apply the projection method to accelerate the optimization process each round, which is $\bigtriangleup \mathcal{M}^{u*} = \left \|  \bigtriangleup \mathcal{M}^{t} \right \| \frac{\bigtriangleup \mathcal{M}^{u}}{\left \|\bigtriangleup \mathcal{M}^{u}  \right\| }$.
where $\bigtriangleup \mathcal{M}^{u*}$ denotes the updated projected model values, $\left \|\cdot  \right \| $ represents the magnitude of the model update 
and $\mathcal{M}^{t}$ corresponds to the global model update value at the $t$-$th$ round.
We proceed with updating the model and have $\mathcal{M}^u = \mathcal{M}^u + \bigtriangleup \mathcal{M}^{u*}$. After training, we obtain the recovery model defined as  $\mathcal{M}^r$. 



\subsection{Implement in FL}
This paper explores the federated training of large models. In this context, we employ Lora~\cite{Lora} fine-tuning to facilitate federated training. As the first step, we apply Lora processing to the model and have  $\{\mathcal{M}_{fix}, \mathcal{M}_{lora}\} \gets Lora(\mathcal{M})$.

For the $t$-$th$ round, the server $\mathcal{S}$ transmits the global model $ \{\mathcal{M}_{fix}, \mathcal{M}_{lora}\}^t$ to the client $\mathcal{N}$, i.e., $ \{\mathcal{M}_{fix}, \mathcal{M}_{lora}\}^{t}_n \gets  \{\mathcal{M}_{fix}, \mathcal{M}_{lora}\}^t$.

Upon receiving the model, the client $\mathcal{N}$ initiates local training using loss function, which is
\begin{equation}
\begin{aligned}
     &\min_{\mathcal{M}\in \mathbb{R}^d} L(\{\mathcal{M}_{fix}, \mathcal{M}_{lora}\}) = \frac{1}{n_i}\sum_{k\in D_i}\ell (\{\mathcal{M}_{fix}, \mathcal{M}_{lora}\})\\
\end{aligned}
\end{equation}
where $L(\{\mathcal{M}_{fix}, \mathcal{M}_{lora}\})$ denotes the loss function employed during local training.

\begin{figure*}
    \centering
    \includegraphics[width=\linewidth]{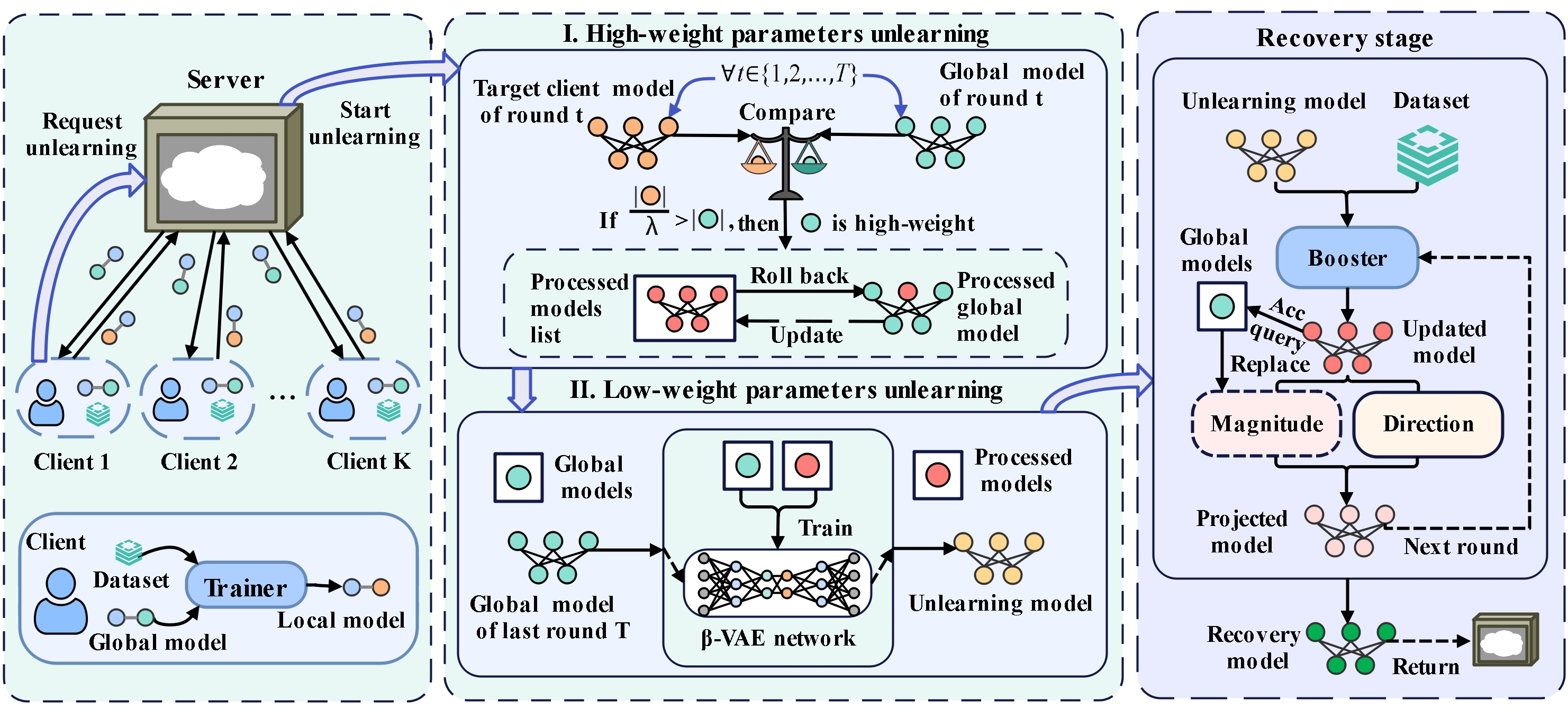}
    \caption{\textbf{The overview of DPUL.}}
    \label{fig:Flow}
\vspace{-0.2cm}
\end{figure*}

The model update process is defined as 
\begin{equation}
\begin{aligned}
    &\bigtriangleup \{\mathcal{M}_{fix}, \mathcal{M}_{lora}\} = \eta \bigtriangledown L(\{\mathcal{M}_{fix}, \mathcal{M}_{lora}\})\\
    &\{\mathcal{M}_{fix}, \mathcal{M}_{lora}\}^* = \{\mathcal{M}_{fix}, \mathcal{M}_{lora}\} - \bigtriangleup \{\mathcal{M}_{fix}, \mathcal{M}_{lora}\}
\end{aligned}
\end{equation}
where $\bigtriangledown L(\{\mathcal{M}_{fix}, \mathcal{M}_{lora}\})$ represents the gradient of the loss function $L(\{\mathcal{M}_{fix}, \mathcal{M}_{lora}\})$. $\eta$ denotes the learning rate, $\{\mathcal{M}_{fix}, \mathcal{M}_{lora}\}^*$ corresponds to the updated local model.

After completing local model training, the client transmits the updated model parameters to the server for aggregation. The aggregation process formula is
\begin{equation}
    \{\mathcal{M}_{fix}, \mathcal{M}_{lora}\}^{t + 1} = \{\mathcal{M}_{fix}, \mathcal{M}_{lora}\}^t + p_i  \sum_{i = 1}^{N} \bigtriangleup \mathcal{M}_{lora}
\end{equation}
where $\bigtriangleup \mathcal{M}_{lora}$ represents the cumulative updates applied to the Lora component of the client model..

Since the $\mathcal{M}_{fix}$ component remains fixed and unchanged, the server only needs to store the updates for $\mathcal{M}_{lora}$, significantly reducing memory consumption.

When client $k$ submits an unlearning request, the server $\mathcal{S}$ first processes $\mathcal{M}_{lora}$ to generate the modified model $\mathcal{M}_{lora}^{'}$, which eliminates the high-weight contributions from the target client using Algorithm \ref{Memory regression}. Next, the server trains the unlearning network $\mathcal{V}$ on the dataset $D = \{(x_i = M_{lora}, y_i = M_{lora}^{'})\}$ and executes the unlearning procedure via Algorithm \ref{Deep unlearning}. As a result, we obtain the unlearning model  $\mathcal{M}_{lora}^u$, forming the final unlearning model $\{\mathcal{M}_{fix}, \mathcal{M}_{lora}^u\}$. Lastly, we enhance the model through projection boosting 
, leading to the recovery model $\{\mathcal{M}_{fix}, \mathcal{M}_{lora}^r\}$. The entire process is visually represented in Figure \ref{fig:Flow}.

\section{Evaluations}

\subsection{Experimental Settings}
\subsubsection{Dataset Description}
We conduct experiments on the following four  datasets: CIFAR-10~\cite{abouelnaga2016cifar}, CINIC-10~\cite{cinic10}, CIFAR-100~\cite{CIFAR100}, and ImageNet-tiny~\cite{ImageNet-Tiny}. The Cifar-10 dataset consists of a training set of 50,000 images and a testing set of 10,000 images. It comprises 10 categories, each containing an equal number of images with a resolution of 32×32 pixels. The CINIC-10 dataset includes a training set of 90,000, a testing set of 90,000, and a validation set of 90,000 images. It consists of 10 categories, each containing an equal number of 32×32 pixel images. The Cifar-100 dataset consists of a training set with 50,000 images and a testing set with 10,000 images. It contains 100 categories, each comprising an equal number of images with a resolution of 32×32 pixels. The ImageNet-tiny dataset comprises a training set with 100,000 images, a testing set with 10,000 images, and a validation set with 10,000 images. It spans 200 categories, each containing an equal number of images at a resolution of 64×64 pixels.

The client dataset follows a non-IID distribution. We employ the Dirichlet distribution to model this, assigning an alpha weight of 1 to each client.
\subsubsection{Model Architectures}
We utilize four large models across different datasets to evaluate the performance of the proposed DPUL. For the CIFAR-10 dataset, we implement the small-sized Vision Transformer (ViTs) model~\cite{Vit}. For the CINIC-10 dataset, we deploy the base-sized Distilled Data-Efficient Image Transformer (DeiTb) model~\cite{Deit}. For the CIFAR-100 dataset, we utilize the base-sized Vision Transformer (ViTb) model. For the ImageNet-tiny dataset, we employ the large-sized Vision Transformer (ViTl) model. These models are pre-trained and undergo Lora processing on the server before being sent to the client for fine-tuning.

For the reconstruction model network, we utilize $\beta$-VAE. The input involves expanding the model into a one-dimensional array, while the output remains in the same format. Ultimately, the output is restored to match the model’s original dimensions.
\subsubsection{Hyperparameters}
In our experiment, we configure the number of FL clients to 10 by default. The training epoch for each client is set to 5, defining the duration of training per communication round with the server. For ViTs and DeiTb, the learning rate $\eta$ is set to $5\times10^{-5}$, while for ViTb and ViTl, it is set to $5\times10^{-3}$. For memory rollback, we assign $\lambda$ a value of 6. In VAE training, we set the learning rate to 0.1 and $\beta$ to 0.5. We deploy the SGD optimization method for local training and VAE training. We set the batch size to 128. For model aggregation, we utilize the FedAvg~\cite{Fedavg} method.
\begin{figure*}
    \centering
    \includegraphics[width=\linewidth]{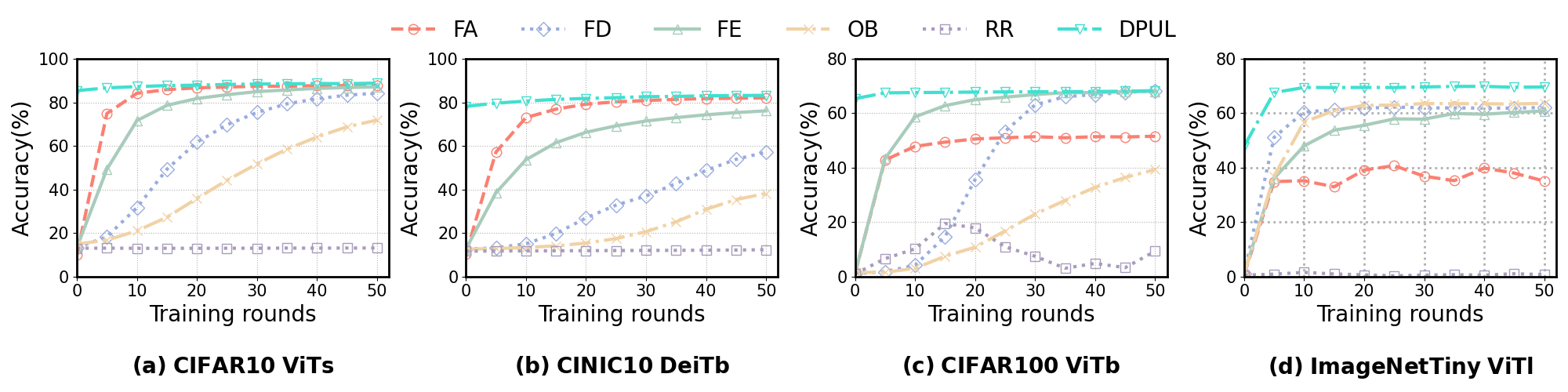}
    \caption{Accuracy recovery of four datasets under different comparison algorithms}
    \label{fig:acc}
\end{figure*}

\begin{figure*}
    \centering
    \includegraphics[width=\linewidth]{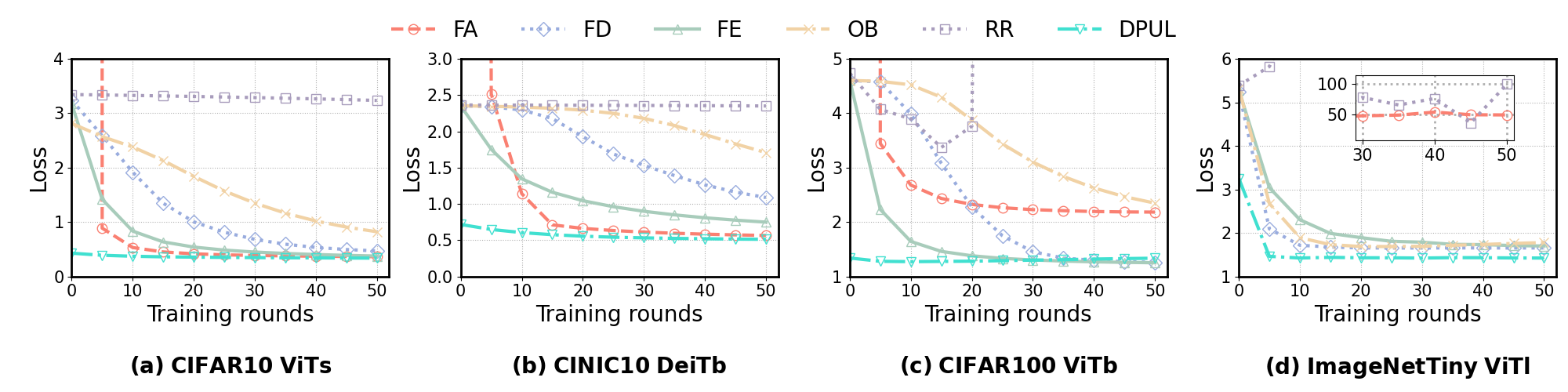}
    \caption{Loss recovery of four datasets under different comparison algorithms}
    \label{fig:loss}
\end{figure*}

\subsubsection{Comparison Methods}
 We compare our algorithm against various methods as follows:
 
 Federated Learning(FL): 
 FL performance is a benchmark for evaluating the effectiveness of different approaches.

 Retrain(RT): RT acts as the baseline for FUL. We aim to expedite this process while ensuring that model recovery performance remains comparable to this baseline.

 Rapid Retrain(RR): RR accelerates RT by leveraging second-order optimization to enhance convergence speed, thereby reducing local training time. In our experiment, we set the number of local training rounds for RR to 1 to further accelerate RT.

 Federated Eraser(FE): FE employs projection techniques to speed up RT by utilizing global model parameters to minimize local training time. In our experiment, we configure the number of local training rounds to 1 and apply global model parameters to local projections for enhanced recovery.
 
 Federated Gradient Ascent(FA): FA leverages unlearning client data for gradient-based boosting during the unlearning process and utilizes additional datasets to enhance recovery performance.

 Federated Knowledge Distillation(FD): FD employs unlabeled datasets to distill knowledge on the server side, facilitating the unlearning process.
 
 Boost training(BT): BT utilizes server-side datasets to aid model recovery~\cite{BT}.

\subsubsection{Backdoor Attack}
We employ backdoor attacks~\cite{backdoor} to assess the effectiveness of the unlearning model. Specifically, we embed feature pixels into the target client data and modify their labels as triggers, while leaving the data from other clients unchanged. When the FL model encounters images with triggers, it classifies them into predefined categories. Conversely, if the unlearning model correctly identifies these images and assigns them to their original classes, we consider the contribution of the unlearning client to be entirely erased. For the CIFAR-10 and CINIC-10 datasets, we insert trigger pixels into images and modify their labels to ‘truck’, causing them to be misclassified as ‘truck’. Similarly, for the CIFAR-100 dataset, trigger images are misclassified as ‘worm’, while those in the ImageNet-Tiny dataset are predicted as 'drumstick'.
\subsubsection{Evaluation Metrics}
We assess the performance of unlearning models within machine learning, considering key metrics such as accuracy, loss value, and computational time.

\subsection{Results}
\subsubsection{Performance of DPUL}
To evaluate the performance of the proposed DPUL method, we compare four datasets and four network models, analyzing the variations in accuracy as the number of recovery rounds increases. Figure \ref{fig:acc} presents the proposed DPUL method alongside five comparison algorithms. 
Our method outperforms other state-of-the-art baselines by approximately 1\% to 5\%.
For instance, in CIFAR-10 at the 50th round, DPUL achieves an accuracy of 88.82\%, which outperforms FA (87.79\%), FD (84.25\%), FE (85.12\%), OB (71.85\%), and RR (13.04\%) by margins of 1.03\%, 4.57\%, 3.70\%, 16.97\%, and 75.78\%, respectively.
Our method exhibits greater stability than FA, FD, and OB while achieving the fastest recovery effects. FA performs significantly worse on the CIFAR100 and ImageNetTiny datasets, with DPUL surpassing it by 16.77\% and 34.51\%, respectively. OB shows weaker performance on CIFAR10 and CINIC10, where DPUL achieves 16.97\% and 45.09\% higher accuracy, respectively. Similarly, FD struggles significantly with accuracy recovery on CINIC10, with DPUL outperforming it by 26.09\%. DPUL method demonstrates a clear advantage on the ImageNetTiny dataset, surpassing all other methods by at least 5\%.

Additionally, we conduct corresponding experiments to analyze changes in loss values, as illustrated in Figure \ref{fig:loss}. The DPUL method exhibits the fastest reduction in loss values, maintaining remarkable stability across all four datasets. Specifically, on the ImageNetTiny dataset, both FA and RR methods show an unexpected increase in loss rather than a decrease, highlighting DPUL’s considerable stability.

\begin{figure*}
    \centering
    \includegraphics[width=\linewidth]{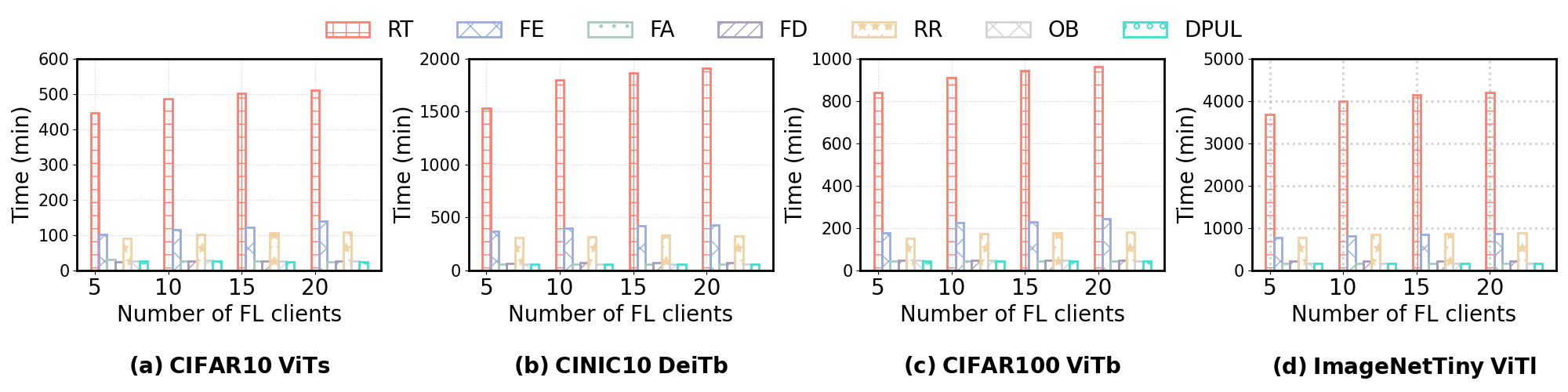}
    \caption{Time consumption of four datasets under different comparison algorithms}
    \label{fig:time}
\end{figure*}

Furthermore, we analyze the variations in runtime for the comparative algorithms with different numbers of clients, as shown in Figure \ref{fig:time}. 
Our method remains independent of the number of clients, ensuring consistent runtime stability.
In contrast, RT, FE, and RR methods exhibit an upward trend in runtime as the number of clients increases, primarily due to the growing number of retraining clients. Notably, DPUL achieves execution speeds nearly 12 times faster than the Retrain method and approximately 4 times faster than the FE and RR methods.

\subsubsection{Validity Verification and Ablation Experiments}
\begin{figure*}
    \centering
    \includegraphics[width=\linewidth]{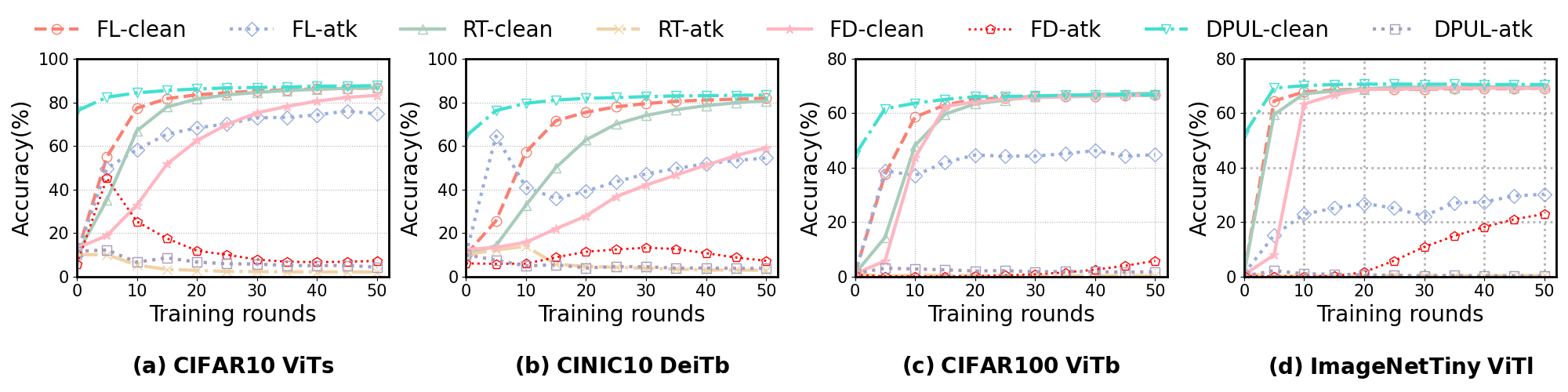}
    \caption{Client contribution during the recovery process}
    \label{fig:atk_acc}
\end{figure*}

Moreover, we simulate the data of the target client using the attack dataset and use backdoor attack accuracy to represent the client’s data contribution. Simultaneously, we evaluate the degree of unlearning our proposed DPUL method achieves. For comparison, we consider FL, RT, and FD, as depicted in Figure \ref{fig:atk_acc}, where the “clean” suffix represents the clean dataset and the “atk” suffix denotes the backdoor attack dataset. We set the number of clients to 5.
From the graph, the Atk accuracy of DPUL matches that of RT, confirming that our method effectively unlearns the client’s contribution. Compared to the FD method operating on the same server, DPUL demonstrates exceptional stability and safety. For instance, in the initial rounds of CIFAR-10, the ATK value of FD exhibits an increasing trend. A similar increase occurs in the last thirty rounds of ImageNetTiny, FD partially restoring attack accuracy. In contrast, DPUL maintains a consistently stable atk value without any upward trend, indicating its superior reliability and safety in practical applications.

Additionally, we conduct ablation experiments with a clean dataset as presented in 
Table \ref{table:union_ablation}. 
A check mark indicates the inclusion of a module, where FL represents federated learning, MP denotes memory regression processing, DU corresponds to parameter reconstruction using the deep unlearning method with $\beta$-VAE networks, BT signifies boosting training, and PR stands for acceleration training via projection. Bold values indicate optimal performance, while underlined values denote the worst results.
The table shows that incorporating all five modules~(FL, MP, DU, BT, and PR) yields the best accuracy and loss values. However, relying solely on RT or a combination of RT and PR leads to a decline in accuracy across all four datasets, underscoring the critical role of MP and DU in the DUPL framework. For instance, on the CINIC10 dataset, accuracy drops by 45.09\% and 23.98\%, respectively. Examining the first row~(FL and MP) alongside the second row~(FL, MP, and DU), we note a slight reduction in accuracy. This is because DU undergoes additional parameter reconstruction to validate its effectiveness.
Furthermore, a comparison between the second-to-last row (FL, MP, DU, BT) and the last row~(FL, MP, DU, BT, PR) demonstrates that the PR module accelerates model recovery, yielding an approximate 1\% to 2\% improvement across all four datasets.
\begin{table*}[ht]
\centering
\setlength{\tabcolsep}{4pt} 
\renewcommand{\arraystretch}{1}
\caption{Ablation Experiments}
\resizebox{\textwidth}{!}{%
\begin{tabular}{@{}cccccccccccccccccccccc@{}} 
\Xhline{1.3pt}
\multicolumn{5}{c}{\multirow{3}{*}{Module}}& \multicolumn{8}{c}{Clean dataset impact} & \multicolumn{8}{c}{Attack dataset impact} \\
\cmidrule(lr){6-13} \cmidrule(l){14-21}
\multicolumn{5}{c}{}&
\multicolumn{2}{c}{CIFAR10} & \multicolumn{2}{c}{CINIC10} & \multicolumn{2}{c}{CIFAR100} & \multicolumn{2}{c}{ImageNetTiny} & 
\multicolumn{2}{c}{CIFAR10} & \multicolumn{2}{c}{CINIC10} & \multicolumn{2}{c}{CIFAR100} & \multicolumn{2}{c}{ImageNetTiny} \\
\cmidrule(lr){1-5} \cmidrule(lr){6-7} \cmidrule(lr){8-9} \cmidrule(lr){10-11} \cmidrule(lr){12-13} \cmidrule(lr){14-15} \cmidrule(lr){16-17} \cmidrule(lr){18-19} \cmidrule(l){20-21}
FL & MP & DU & BT & PR & Acc & Loss & Acc & Loss & Acc & Loss & Acc & Loss & Clean & Atk & Clean & Atk & Clean & Atk & Clean & Atk \\
\Xhline{1.3pt}
 $\surd$&  &  &  &  &89.12  &0.32  &83.28  &0.50  &\textbf{69.48}  &\textbf{1.21}  &62.02  &1.61  & 86.62 & \underline{74.66} & 81.90 & \underline{54.46} & 66.65 & \underline{44.72} & 69.12 & \underline{30.09} \\
$\surd$ & $\surd$ &  &  &  & 88.79 & 0.33 & 81.69 & 0.55 & 65.14 & 1.33 & 51.55 & 2.02 & 79.69 & 7.19 & 70.33 & 4.17 & 44.29 & 0.80 & \underline{50.98} & 0.26 \\
$\surd$ & $\surd$ & $\surd$ &  &  & 85.37 & 0.42 & 78.13 & 0.71 & 65.12 & 1.34 & \underline{48.98} & \underline{2.11} & 75.87 & 7.63 & 64.30 & 9.20 & 45.32 & 1.11 & 56.41 & 0.27 \\
 &  &  & $\surd$ &  & \underline{71.85} & \underline{0.82} & \underline{38.12} & \underline{1.71} & \underline{39.29} & \underline{2.34} & 63.67 & 1.78 &\underline{71.83}  &4.54  &\underline{38.56}  &12.51  &\underline{39.34}  &\textbf{0.32}  &63.89  &\textbf{0.13}  \\
 &  &  & $\surd$ & $\surd$ & 80.51 & 0.58 & 59.23 & 1.20 & 52.45 & 1.89 & 66.66 & 1.64 &80.79  &\textbf{3.02}  &59.27  &4.69  &52.36  &0.51  &66.88  &0.26  \\
$\surd$ &  &  & $\surd$ &  &\textbf{89.39}  &\textbf{0.31}  &\textbf{83.61}  &\textbf{0.49}  &68.10  &1.29  &69.27  & 1.46 &\textbf{87.89}  & 46.12 & 83.19 & 23.86 & \textbf{66.98} & 19.40 & 69.41 & 5.28 \\
$\surd$ & $\surd$ & $\surd$ & $\surd$ &  & 88.13 & 0.35 & 81.98 & 0.57 & 67.01 & 1.41 & 69.01 & 1.51 &87.01  &5.01  &82.56  &4.15  &65.98  &1.98  &69.44  &0.63  \\
$\surd$ & $\surd$ & $\surd$ & $\surd$ & $\surd$ & 88.82 & 0.33& 83.21 & 0.51 & 68.12 & 1.29 & \textbf{69.92} & \textbf{1.47} & 87.65 & 4.18 & \textbf{83.34} & \textbf{3.71} & 66.63 & 1.67 & \textbf{70.12} & 0.58 \\
\Xhline{1.3pt}
\end{tabular}%
}
\label{table:union_ablation}
\end{table*}

Furthermore, we perform ablation experiments on five clients with the attack dataset, where the target client contributes backdoor attack datasets, to evaluate the role of each module in contribution unlearning, as presented in the same Table \ref{table:union_ablation}.
Backdoor attacks are injected into the target client within the FL framework. The first row of FL represents the accuracy of the clean dataset for federated training alongside the backdoor attack dataset. At this stage, the backdoor attack accuracy is the baseline for unlearning and represents the maximum value.
Upon introducing the MP module, we observe that atk accuracy is vastly reduced, reaching optimal values of 0.8\% and 0.26\% in the CIFAR100 and ImageNetTiny datasets, respectively, which confirms that our proposed memory rollback module effectively unlearns the client’s primary contribution.
When only Boost and Projection are applied, the backdoor attack accuracy somewhat decreases. However, it remains relatively high, indicating that Boost and Projection alone cannot effectively unlearn the client’s contribution. For instance, in the CIFAR10 dataset, attack accuracy still stands at 46.12\%. 
For residual minor contributions, incorporating the DU module (fourth row) decreases clean accuracy, whereas atk accuracy remains relatively unchanged, demonstrating its role in further unlearning. In the final boosting phase, attack accuracy does not recover, verifying that the client’s contribution is not reinstated during Boost training. For instance, in the CIFAR10 and CINIC10 datasets, the lowest atk accuracy values of 4.18\% and 3.17\% confirm that our proposed DPUL unlearns the client’s contribution.

\subsubsection{Performance under Different FL Conditions}
\begin{figure*}
    \centering
    \includegraphics[width=\linewidth]{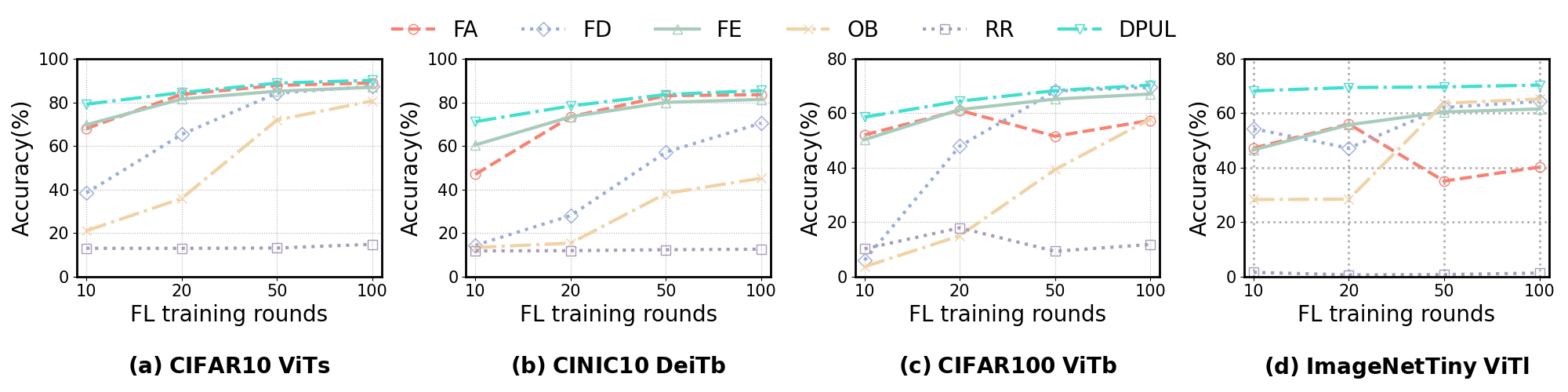}
    \caption{The influence of different rounds of FL on experiments}
    \label{fig:Epoch}
\end{figure*}

Moreover, we evaluate the performance of our method across different FL training rounds and assess its recovery over the same number of rounds, as illustrated in Figure \ref{fig:Epoch}. The figure demonstrates that our DPUL method consistently outperforms the compared algorithms, particularly in scenarios with fewer training rounds. 
For instance, during 10 rounds of FL training and recovery on CIFAR10, our DPUL method surpasses FA by 11.17\%, OB by 58.21\%, FD by 45.29\%, FE by 9.46\%, and RR by 66.24\%.
As the number of rounds increases, our method continues to improve while maintaining a performance advantage over the comparative approaches. 
For example, in CIFAR100, after 100 rounds of FL training, unlearning, and recovery, our DPUL method outperforms FA by 13\%, FD by 0.8\%, FE by 3.28\%, OB by 12.5\%, and RR by 58.61\%.
In ImageNetTiny, our method exhibits minimal variation with increasing rounds, which is attributable to pre-training on the ImageNet-21k dataset, ensuring effective fine-tuning on ImageNetTiny. Additionally, ImageNetTiny is a relatively large dataset, and both DPUL and the comparison algorithms utilize a larger sample size than the first three datasets, contributing to its superior recovery performance.

\begin{figure*}
    \centering
    \includegraphics[width=\linewidth]{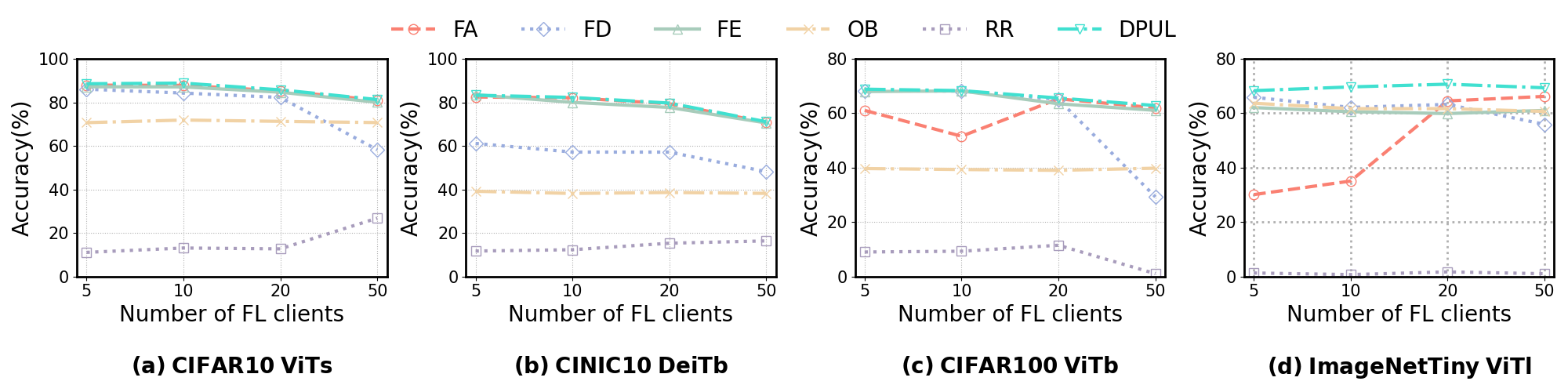}
    \caption{The impact of different numbers of clients on experiments}
    \label{fig:Num}
\vspace{-0.3cm}    
\end{figure*}

Additionally, we analyze the performance of our DPUL method under varying numbers of clients during FL, as presented in Figure \ref{fig:Num}.
From the graph, we observe that as the number of clients increases, the comparison algorithms and DPUL exhibit a downward trend, which is affected by FL heterogeneity. In the CIFAR100 and ImageNetTiny datasets, the expansion in classification categories leads to varying degrees of performance degradation in comparative algorithms such as FA and FD due to changes in client numbers, whereas DPUL maintains stable performance.
At 50 clients, FD experiences a particularly sharp decline, dropping to 29.25 in CIFAR100, which is 33.57\% lower than DPUL. In contrast, DPUL’s performance on ImageNetTiny remains largely unaffected by the increasing number of clients. 

\subsubsection{The Influence of Hyperparameters}
Moreover, we analyze the impact of hyperparameters, as shown in 
Table \ref{Table:union_hyperparameters}.

We first analyze the impact of the VAE loss training coefficient $\beta$ value in Equation \ref{beta}.
We select three values for this experiment: 0.1, 0.5, and 1.5.
The table shows that when $\beta = 0.1$, occurrences of the worst performance are more frequent. In contrast, selecting 0.5 yields the highest number of best-performing cases and the fewest worst-performing occurrences, suggesting a lower $\beta$ provides high fidelity. When $\beta = 1.5$, both best and worst performances appear less frequently than 0.5, suggesting that a higher $\beta$ provides high reconstruction ability.



Additionally, we analyze the impact of high-weight coefficient $\lambda$ values in the memory rollback method.
We select three values of 1, 6, and 10 for the experiment. From the table, we observe that when the $\lambda$ value is 1, DPUL exhibits the worst performance, with clean accuracy being the lowest across all four datasets, indicating that a lower $\lambda$ tends to favor preserving model accuracy. When the $\lambda$ value is set to 10, although clean accuracy reaches its highest level, atk accuracy declines significantly, indicating that a higher $\lambda$ value enforces stronger removal of client-specific contributions. The experiment shows that a $\lambda$ value is better set around 6, achieving a balance between preserving clean accuracy and effectively removing client-specific information.  



Finally, we analyze the impact of slice quantity I in Equation \ref{slice}. We record the accuracy of both clean and backdoor attack datasets
to assess their specific effects. 
When the I value is set to 1, the accuracy of both clean and atk datasets frequently ranks among the worst, with the clean dataset accuracy being the lowest across all four datasets, suggesting a single VAE model yields poor results. 
As the number of I increases, clean dataset accuracy follows an upward trend, and atk accuracy fluctuates randomly, which indicates that increasing the number of slices consistently improves the performance.

\begin{table}[htbp]
\centering
\setlength{\tabcolsep}{4pt} 
\renewcommand{\arraystretch}{1}
\caption{Combined Analysis of Hyperparameter Effects}
\resizebox{0.45\textwidth}{!}{%
\begin{tabular}{@{}cccccccccccc@{}}
\Xhline{1.3pt}
\multicolumn{2}{c}{\multirow{3}{*}{Dataset}}& \multicolumn{3}{c}{$\beta$} & \multicolumn{3}{c}{$\lambda$} & \multicolumn{4}{c}{I} \\
\cmidrule(lr){3-5} \cmidrule(lr){6-8} \cmidrule(l){9-12}
 &  & 0.1 & 0.5 & 1.5 & 1 & 6 & 10 & 1 & 3 & 5 & 10 \\
\Xhline{1.3pt}
\multirow{2}{*}{CIFAR10} 
& Clean & \underline{75.3} & 75.71 & \textbf{76.28} & \underline{13.32} & 75.71 & \textbf{80.82} & \underline{47.82} & 74.3 & 75.56 & \textbf{75.71} \\
& Atk   & \underline{10.57} & \textbf{10.22} & 10.41 & 12.45 & \textbf{10.22} & \underline{36.05} & \underline{15.3} & 14.87 & \textbf{8.11} & 10.22 \\

\multirow{2}{*}{CINIC10} 
& Clean & \underline{64.67} & \textbf{65.46} & 65.14 & \underline{12.71} & 65.46 & \textbf{69.15} & \underline{13.39} & 55.16 & 54.1 & \textbf{65.46} \\
& Atk   & \underline{10.82} & \textbf{9.74} & 10.48 & \textbf{5.91} & 9.74 & \underline{41.86} & \textbf{7.27} & 9.79 & \underline{11.07} & 9.74 \\

\multirow{2}{*}{CIFAR100} 
& Clean & \textbf{45.28} & 45.16 & \underline{45.05} & \underline{1.06} & 45.16 & \textbf{54.37} & \underline{42.05} & \textbf{45.74} & 44.58 & 45.16 \\
& Atk   & 1.13 & \underline{1.31} & \textbf{1.07} & \textbf{0.57} & \underline{1.31} & \underline{16.65} & 1.17 & \textbf{1.101} & 1.17 & \underline{1.31} \\

\multirow{2}{*}{ImageNetTiny} 
& Clean & 56.37 & \textbf{56.52} & \underline{56.06} & \underline{2.92} & 56.52 & \textbf{59.9} & \underline{32.27} & 53.93 & 56.23 & \textbf{56.52} \\
& Atk   & \textbf{0.37} & \underline{0.46} & \textbf{0.37} & \textbf{0.27} & 0.46 & \underline{1.94} & \textbf{0.19} & \underline{0.61} & 0.37 & 0.46 \\
\Xhline{1.3pt}
\end{tabular}%
}
\label{Table:union_hyperparameters}
\end{table}

\section{Conclusion}
This paper introduces DPUL, a novel method that first leverages weight-aware parameters for server-side unlearning to avoid the privacy pitfalls that appeared in previous server-side unlearning methods. The approach begins with memory rollback, which filters and eliminates high-weight parameter contributions of the target client. For low-weight parameter contributions, we employ a $\beta$-VAE network to reconstruct and facilitate unlearning.
Moreover, we employ a multi-head training mechanism by partitioning the VAE model parameters into multiple segments and training in parallel for better performance. 
The experimental results reveal that our DPUL method achieves higher recovery accuracy, accelerated performance, and stability. The source code and datasets are made available at \href{https://github.com/00taotao/DPUL}
{https://github.com/00taotao/DPUL}.




\bibliographystyle{IEEEtran}
\bibliography{references}

\end{document}